\documentclass[letterpaper, 10 pt, conference]{ieeeconf}  

\IEEEoverridecommandlockouts                              

\usepackage{times}
\usepackage{parskip}
\usepackage[labelfont=bf,textfont=it]{caption}

\usepackage{graphicx}        
\usepackage{url}
\usepackage{hyperref}
\usepackage{amsmath,amssymb}
\usepackage{multirow}
\usepackage[table,xcdraw]{xcolor}
\usepackage{subfigure}
\usepackage{caption}
\usepackage[ruled, vlined, linesnumbered]{algorithm2e}
\usepackage{float}
\usepackage{color}
\usepackage{bm}
\usepackage{textcomp}

\newcommand{\norm}[1]{\lVert#1\rVert}
\newcommand{\M}[1]{$\mathcal{M}$}

\title{\LARGE \bf
Pedestrian Dominance Modeling for Socially-Aware Robot Navigation}

\author{Tanmay Randhavane$^{1}$, Aniket Bera$^{1}$, Emily Kubin$^{2}$, Austin Wang$^{1}$, Kurt Gray$^{3}$, and Dinesh Manocha$^{4}$
\thanks{$^{1}$Authors from the Department of Computer Science, University of North Carolina at Chapel Hill, USA}
\thanks{$^{2}$Author from the Department of Psychology, Tilburg University, North Brabant, Netherlands}
\thanks{$^{3}$Author from the Department of Psychology and Neuroscience, University of North Carolina at Chapel Hill, USA}
\thanks{$^{4}$Author from the Department of Computer Science and Electrical \& Computer Engineering
, University of Maryland at College Park, USA
}%
}

\begin{document}

\maketitle
\thispagestyle{empty}
\pagestyle{empty}

\begin{abstract}
We present a Pedestrian Dominance Model (PDM) to identify the dominance characteristics of pedestrians for robot navigation. Through a perception study on a simulated dataset of pedestrians, PDM models the perceived dominance levels of pedestrians with varying motion behaviors corresponding to trajectory, speed, and personal space. At runtime, we use PDM to identify the dominance levels of pedestrians to facilitate socially-aware navigation for the robots. PDM can predict dominance levels from trajectories with \texttildelow 85\% accuracy. Prior studies in psychology literature indicate that when interacting with humans, people are more comfortable around people that exhibit complementary movement behaviors. Our algorithm leverages this by enabling the robots to exhibit complementing responses to pedestrian dominance. We also present an application of PDM for generating dominance-based collision-avoidance behaviors in the navigation of autonomous vehicles among pedestrians. We demonstrate the benefits of our algorithm for robots navigating among tens of pedestrians in simulated environments.



\end{abstract}

\section{Introduction}
When humans navigate through crowds, they explicitly or implicitly estimate information related to the trajectories and velocities of other people. Additionally, they also try to estimate other social information such as emotions and moods, which helps them predict physical characteristics related to motion and navigation. Someone who looks upset may have more erratic motion paths and someone who looks preoccupied may be less aware of approaching obstacles. 

\begin{figure}
  \centering
  \includegraphics[width =1.0 \linewidth]{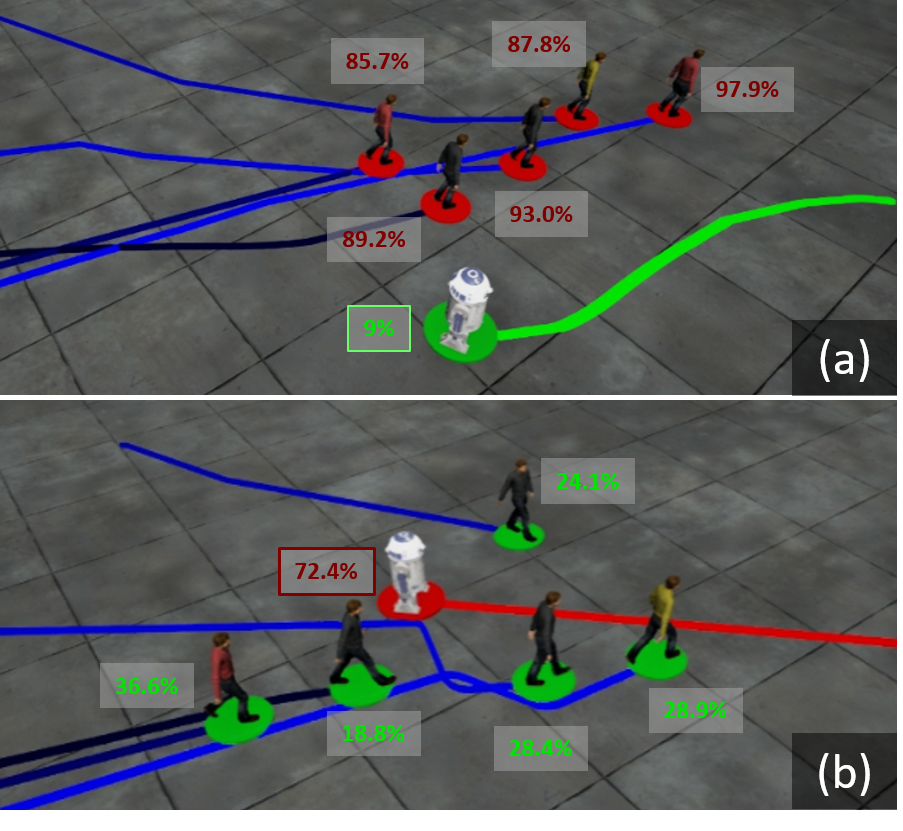}
    \vspace*{-15pt}
  \caption{The robot performs socially-aware navigation among pedestrians. Robot identifies the dominance characteristics of pedestrians based on their trajectories (marked in \textbf{\textcolor{blue}{blue}}) and exhibits complementary behavior. (a) Pedestrians are identified as dominant and the robot therefore exhibits submissive characteristics (marked in \textbf{\textcolor{green}{green}}) and turns away from those pedestrians. (b) Pedestrians are identified as submissive and the robot therefore exhibits dominant characteristics (marked in \textbf{\textcolor{red}{red}}) and expects the pedestrians to move out of its way. We show the identified dominance values for all the pedestrians and the complementary values for the socially-aware robot. For full video, refer to the supplementary material.}
  \label{fig:dominanceValues}
    \vspace*{-20pt}
\end{figure}

Considerable literature in motion planning and human-robot interaction has incorporated social information about humans. Techniques have been proposed for socially-aware robot navigation that predict the movement or actions of human pedestrians~\cite{kruse2013human,okal2016learning}. These algorithms can generate paths that consider the right of way, personal spaces, and other social norms.

Robots may not have social perception abilities as rich as humans, but with their current perception capabilities (e.g., cameras, depth sensors, etc.), they can identify certain personality traits or emotions of humans~\cite{bera2017sociosense}. Research has shown that trajectories, facial expressions, and appearance can be used to automatically assess human personalities, emotions, and moods~\cite{todorov2015social,randhavane2017f2fcrowds}.

Humans make use of personality characteristics such as \textit{psychological dominance}~\cite{ridgeway1987nonverbal} to predict how individuals will act in different situations. Dominance is an essential psychological characteristic in all social organisms including humans. In general, people at the top of social hierarchies (i.e., people who have power, influence, or higher social standing) are \textit{dominant}, which means they tend to act more aggressively~\cite{mcdermott2017dominance} and expect others to accommodate and acquiesce to their behavior. Dominance is important for pedestrian navigation because when dominant pedestrians walk among others they expect others to move out of their way, while submissive pedestrians do not have this expectation~\cite{ridgeway1987nonverbal}. Humans respond to dominant behavior in two ways; they either mimic the behavior (respond in a dominant manner) or they complement it (respond in a submissive manner). Research shows that humans who exhibit complementing responses (dominance in response to submission and submission in response to dominance) are more comfortable with their interacting partner and like them more~\cite{tiedens2003power}. 

Though social hierarchies are unclear in large crowds, psychological research reveals that all individuals generally feel more dominant or less dominant relative to others, and this trait can be estimated from their movements. As robots move through human environments, being able to assess the dominance levels of humans gives them a better ability to navigate through crowds. In particular, they will be able to complement the behaviors of the human pedestrians and, as a result, create more positive relationships with humans~\cite{tiedens2003power}.


To assess dominance, humans process many social cues, including posture \cite{lukaszewski2016role}, gait \cite{montepare1987identification}, and facial expressions~\cite{todorov2015social}. Computer vision and deep learning techniques can be used to identify these cues, though their accuracy can vary.

\begin{figure*}[t]
  \centering
  \includegraphics[width =1.0\linewidth]{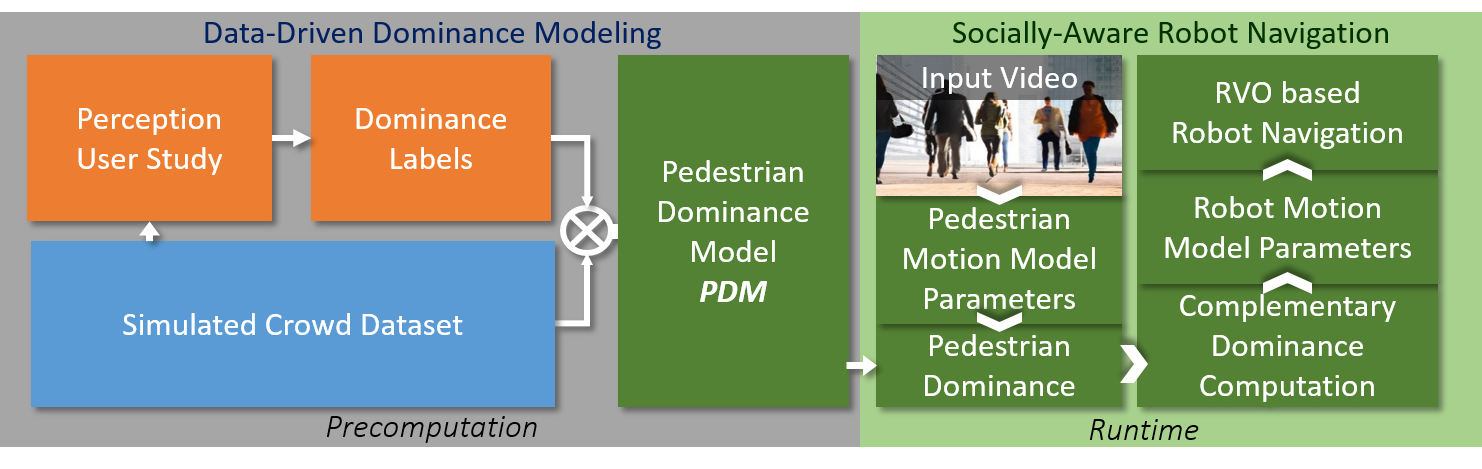}
     \vspace*{-15pt}
\caption{\textbf{Overview}: We provide an overview of our dominance modeling and socially-aware robot navigation that accounts for pedestrian dominance. We present a novel Pedestrian Dominance Model \textit{(PDM)} that models perceived dominance levels using a simulated dataset of pedestrians with varying motion model parameters. At runtime, PDM is used to identify the dominance levels of pedestrians in a streaming input video. We then use PDM to design a socially-aware navigation algorithm such that the robot exhibits dominance behaviors that are complementary to the pedestrians' dominance behaviors.}
    \vspace*{-15pt}
  \label{fig:overview}
\end{figure*}

\textbf{Main Results}:
In this paper, we focus on using the trajectories of and proximities among people to model dominance characteristics and then use them to guide a robot's navigation. Humans often express and perceive personality traits through movements, and we use that information to identify their dominance levels. We present a novel algorithm that uses trajectory information to assess pedestrians' dominance levels and then uses these assessments to facilitate socially-aware robot navigation around humans. Figure~\ref{fig:overview} presents an overview of our algorithm for computing the psychological dominance of pedestrians in real-time. Given a video stream as an input, we extract trajectories of all pedestrians and estimate motion model parameters for each pedestrian using Bayesian learning algorithms. Based on the motion model parameters of a pedestrian, our novel data-driven pedestrian dominance model \textit{(PDM)} computes the psychological dominance of the pedestrian.


We generate a simulated dataset of pedestrians with varying dominance levels by adjusting the motion model parameters. Using a perception study, we obtain dominance values for pedestrians in this dataset. Our pedestrian dominance model (PDM) establishes a linear mapping between the motion model parameters and pedestrian dominance using multiple linear regression. PDM can predict dominance levels from motion model parameters with \texttildelow 85\% accuracy.

We design a dominance-aware navigation algorithm that takes pedestrian dominance into account during runtime. A key contribution of our algorithm is facilitating the robots in exhibiting complementary behaviors to the dominance of pedestrians in addition to providing collision-free navigation.
We also apply pedestrian dominance to navigate an autonomous vehicle among pedestrians using dominance-based decisions. 

Some of the main benefits of our approach include: 

\noindent {\bf 1. Robust:} Our algorithm is general and can account for noise in the extracted pedestrian trajectories. Our algorithm can identify the dominance level of each pedestrian in scenarios corresponding to low- and medium- density crowds.

\noindent {\bf 2. Dominance Model:} Our algorithm establishes a mapping between dominance and pedestrian trajectories that can be used to make judgments about a pedestrian's internal state.

\noindent {\bf 3. Socially-Aware Robot Navigation:} Our real-time robot navigation algorithm generates navigation behaviors that are complementary to the dominance levels of pedestrians. Complementarity in dominance relationships has been known to result in more rapport and comfort~\cite{tiedens2003power}.

The rest of the paper is organized as follows. Section II provides an overview of related work in robot navigation. 
In Section III, we present the details of our novel pedestrian dominance model. We present details of the dominance-aware navigation scheme for robots in Section IV. Section V provides an application for navigating autonomous vehicles among pedestrians. 

\section{Related Work}
In this section, we give a brief overview of prior work on robot navigation and behavior models. 

\subsection{Robot Navigation in Human Environments}
Starting with early systems that use robots as tour guides in museums (RHINE~\cite{burgard1998interactive} and MINERVA~\cite{thrun1999minerva}), there has been considerable work on robot navigation in urban environments~\cite{bauer2009autonomous,kummerle2015autonomous}. Potential-based approaches have been used for robot path planning in dynamic environments~\cite{Pradhan11_potential_function}. Pedestrian trajectory prediction using Bayesian velocity obstacles~\cite{kim2014brvo} and a Partially Closed-loop Receding Horizon Control~\cite{DuToit10} has been used for robot navigation around pedestrians. Pedestrian trajectories have also been predicted by learning motion characteristics from real-world trajectories~\cite{Ziebart2009,Henry10_IRL}.

\subsection{Socially-Aware Robot Navigation}
Most of the robot navigation approaches mentioned above can be extended by psychological and social information to perform socially-aware navigation~\cite{pandey2010framework,bera2018socially,bera2018classifying}. Some methods model the cooperation between robots and humans by using probabilistic models~\cite{RobNaviProb_Cafe} and data-driven techniques~\cite{pfeiffer2016predicting,beraglmp}. Learning-based approaches have also been used to perform social navigation that respects social conventions and norms~\cite{chen2017socially,kretzschmar2016socially}. Many approaches explicitly model social constraints~\cite{sisbot2007human,kirby2009companion} or consider interactions and personal space~\cite{barnaud2014proxemics} to enable human-aware navigation. There is work on developing learning algorithms that allow a robot to announce its objective to a human~\cite{huang2017enabling}. Other systems have treated an autonomous (robot) vehicle and a human driver as a dynamical system where the actions of the robot and human affect each other~\cite{sadigh2016planning}.

Researchers are also studying psychological factors (such as personality) that shape body movement~\cite{collier1985emotional}. Models have been proposed ~\cite{ballrelating} and developed~\cite{bera2017sociosense,bera2017ijcai} to predict personality traits and predict future actions based on the body movements of pedestrians. Relationships between personality differences and spatial distances between pedestrians and robots have also been explored~\cite{walters2005influence}. Our work in modeling psychological dominance is complementary to these methods of using personality traits as a predictor of movement and can be combined with most of these methods. 

\subsection{Dominance Modeling}
There has been work on computing dominance in psychology and AI literature~\cite{randhavane2019modeling}. Supervised learning approaches have been proposed to learn dominance based on audio and verbal cues~\cite{rienks2005automatic}. An extension also includes visual activity cues in a multi-camera, multi-microphone setting~\cite{jayagopi2009modeling}. Pair-wise influence in face-to-face interactions has also been modeled based on vocal cues~\cite{otsuka2006quantifying}. Most of this work is based on verbal or vocal cues, whereas we model dominance as a function of trajectory information, which is of interest in robot navigation.



\section{Dominance Learning}
Our goal is to model the levels of perceived dominance from pedestrian trajectories. Guy et al.~\cite{Guy2011Personality} used a simulated dataset of pedestrians to assess the perception of various personality traits and their relation to different low-level motion parameters. We use a similar data-driven approach to model pedestrians' psychological dominance. We present the details of our perception study in this section and derive the Pedestrian Dominance Model (PDM) from the results.

\subsection{Perception User Study}
\subsubsection{Study Goals}
This web-based study aimed to obtain the dominance labels for a simulated dataset of pedestrians. We use a 2-D motion model~\cite{van2011reciprocal} based on reciprocal velocity obstacles (RVO) to model the motion of pedestrians. We obtain scalar values of perceived dominance for different sets of motion model parameters used in modeling pedestrian motion.

\subsubsection{Participants}
We recruited $390$ participants ($217$ male, $172$ female, $1$ other, $\bar{x}_{age}$ = $35.29$, $s_{age}$ = $11.13$) from Amazon MTurk to answer questions about a dataset of simulated videos. 

\subsubsection{Dataset}
Based on prior work~\cite{Guy2011Personality}, we use a dataset with varying sets of motion parameters values. In particular, we consider the following motion parameters for each pedestrian:
\begin{itemize}
\item Neighbor Distance (maximum distance of neighbors affecting the agent),
\item Maximum Neighbors (maximum number of neighbors affecting the local behavior of an agent),
\item Planning Horizon (how far ahead the agent plans),
\item Effective Radius (how far away an agent stays from other agents), and
\item Preferred Speed.
\end{itemize}

We represent these motion model parameters as a vector ($\mathbf{P}\in\mathbb{R}^5$): \textit{Neighbor~Dist, Max~Neighbors, Planning~Horiz, Radius, Pref~Speed}.


\begin{table}[ht]
\centering
\begin{tabular}{|l|c|c|c|}
\hline
\multicolumn{1}{|c|}{Parameter (unit)} & Min & Max & Default \\ \hline
Neighbor Dist (m)                      & 3   & 30  & 15      \\ \hline
Max Neighbors                          & 1   & 40  & 10      \\ \hline
Planning Horiz (s)                     & 1   & 24  & 24      \\ \hline
Radius (m)                             & 0.3 & 2   & 0.8     \\ \hline
Pref Speed (m/s)                       & 1.2 & 2.2 & 1.4     \\ \hline
\end{tabular}
\caption{\textbf{Values of Motion Parameters}: We present the range and default values of motion parameters used to create the simulated dataset. These values cover the range of values observed in the real world.}
\vspace*{-15pt}
\label{tab:dataRange}
\end{table}

In this study, we created simulated videos of four different scenarios (refer to the supplementary video). In each scenario, there is a single highlighted agent wearing a red shirt with a yellow disk drawn beneath him. 
\begin{itemize}
\item Pass-Through \textit{(PT)}: The highlighted agent moves through a cross-flow of $40$ agents.
\item Corridor \textit{(C)}: The highlighted agent and a group of 5 agents start on the opposite ends of a corridor and walk towards the other end. 
\item Standing Group \textit{(SG)}: The highlighted agent navigates a group of 5 standing agents to reach his goal. 
\item Narrow Exit \textit{(NE)}: The highlighted agent exits through an opening along with 30 other agents.
\end{itemize}
 
Each scenario was simulated using a 3D crowd simulation framework~\cite{narangsimulating} that uses RVO~\cite{van2011reciprocal} for collision avoidance. We generated $12$ videos for each scenario by randomly varying the highlighted agent's motion parameters. The non-highlighted agents were assigned the default parameters in all cases. Table~\ref{tab:dataRange} shows the range and the default values of the parameters used. In each case, we also showed a \textit{Reference Video}  side-by-side to the \textit{Question Video} in which all the agents were simulated with the default parameters. We used the same Reference Video for all the videos in each scenario.

\subsubsection{Procedure}
In the web-based study, participants were asked to watch a random subset of $6$ videos from one of the four scenarios. Participants then answered whether the highlighted agent was Submissive, Withdrawn, Dominant, or Confident on a 5-point Likert scale from Strongly Disagree (1) - Strongly Agree (5). Our choice to use these adjectives to represent dominance was based on previous studies from the psychology literature~\cite{dunbar2005perceptions,ridgeway1987nonverbal}. We considered all the adjectives ($12$) used in these studies and decided to use the four adjectives that capture general impressions of an individual's dominance based on a pilot study. Participants were presented the videos in a randomized order and could watch the videos multiple times if they wished. Before completing the study, participants also provided demographic information about their gender and age.


\subsubsection{Analysis}
We average the participant responses to each video to obtain a mean value corresponding to each dominance adjective: $V_{sub}, V_{with}, V_{dom}, V_{conf}$. We obtain a scalar label for dominance $d$ by combining these average values and normalizing to convert them into a range of 0-1:
\begin{eqnarray}
d = \frac{(V_{dom} + V_{conf} + 6 - V_{sub} + 6 - V_{with}) - 4}{16}
\end{eqnarray}

\begin{figure}
  \centering
  \includegraphics[width =1.0 \linewidth]{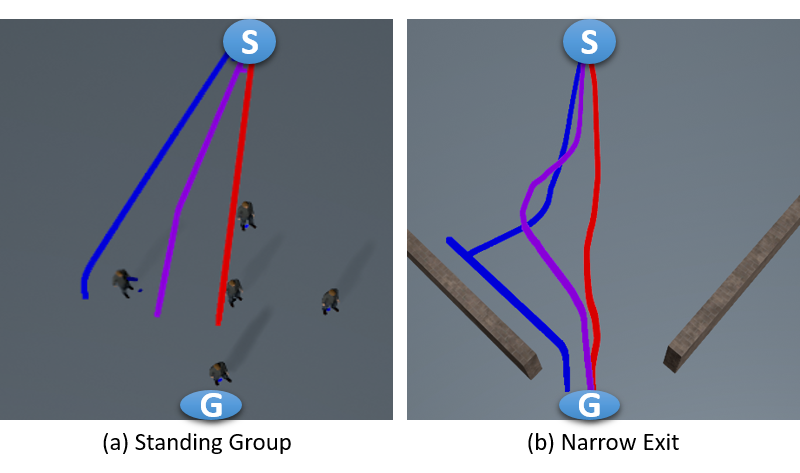}
    \vspace*{-15pt}

  \caption{\textbf{Trajectories}: We present sample trajectories of the highlighted agent in the Standing Group and Narrow Exit scenarios. \textit{S} and \textit{G} represent the start and goal positions of the pedestrian. Dominant pedestrians (\textbf{\textcolor{red}{red}}) take a more direct path and expect others to move out of their way~\cite{ridgeway1987nonverbal}. In the Standing Group scenario, these dominant pedestrians also pass through groups of pedestrians (\textbf{\textcolor{red}{red}} and \textbf{\textcolor{purple}{purple}}). Submissive pedestrians (\textbf{\textcolor{blue}{blue}}) are diverted more easily attempting to avoid others and walk around groups of pedestrians standing in their path.}
  \label{fig:traj}
    \vspace*{-10pt}

\end{figure}

\begin{figure}
  \centering
  \includegraphics[width =1.0 \linewidth]{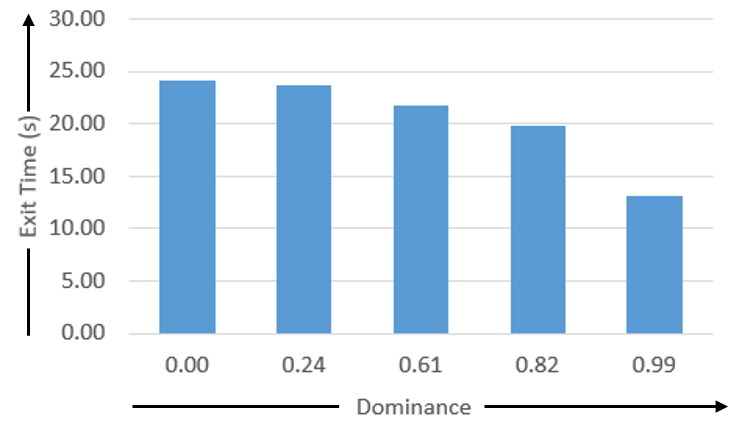}
    \vspace*{-15pt}
  \caption{\textbf{Exit Times}: We provide the time at which the highlighted agent can pass through the narrow exit in the Narrow Exit scenario. Dominant agents take more direct paths and can exit faster than submissive agents.}
  \label{fig:narrowExitTimes}
    \vspace*{-12pt}
\end{figure}

We present the trajectories of the highlighted agent obtained by varying the motion model parameters in the Standing Group and Narrow Exit scenarios (Figure~\ref{fig:traj}). Dominant pedestrians take a more direct path and expect others to move out of their way~\cite{ridgeway1987nonverbal}. In the Standing Group scenario, they also pass through groups of pedestrians, and the pedestrians make way. Submissive pedestrians are diverted more easily in attempts to avoid others and walk around groups of pedestrians standing in their path. We also analyze the time taken by the highlighted agent to pass through the narrow exit in the Narrow Exit scenario in Figure~\ref{fig:narrowExitTimes}. More dominant agents exited the area quicker than submissive agents. As seen in Figure~\ref{fig:narrowExitTimes}, a very dominant agent took 54\% less time than the most submissive agent to exit the area.


\subsection{Pedestrian Dominance Model (PDM)}
Using the perception study, we have obtained the dominance value $d_{i}$ corresponding to each variation of the motion model parameters $\textbf{P}_{i}$. Given these $48$ data points corresponding to $48$ videos in the simulated dataset, we can fit a model for dominance computation using multiple linear regression. Other forms of regressions can also be employed. 

We use the difference between the highlighted agents' parameters in the Question Video and those in the Reference Video as input to the regression. This avoids the computation of an offset in the regression. Like Guy et al.~\cite{Guy2011Personality}, we also normalized the input by dividing the parameters by half of their range to increase the stability of the regression. Thus, our \textit{Pedestrian Dominance Model (PDM)} takes the following form:

\begin{small}
\begin{equation} \label{eq:mapping}
d
  =
  \bf{D}*
\begin{pmatrix}
 \frac{1}{13.5}(Neighbor~Dist - 15)\\[0.2em]
 \frac{1}{19.5}(Max~Neighbors - 10)\\[0.2em]
\frac{1}{11.5}(Planning~Horizon - 24)\\[0.2em]
\frac{1}{0.85}(Radius - 0.8)\\[0.2em]
\frac{1}{0.5}(Pref~Speed - 1.4)\\[0.2em]
\end{pmatrix}
\end{equation}
 \end{small}
    \vspace*{-17pt}
where 
\begin{equation}\label{eq:coeffs}
 \bf{D} = \begin{pmatrix}
0.01 & -0.07 & -0.41 & 0.05 & 0.14
\end{pmatrix}
\end{equation}
\vspace*{-10pt}

\begin{figure}
  \centering
  \includegraphics[width =1.0 \linewidth]{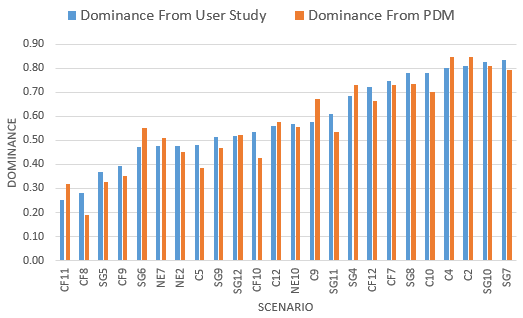}
    \vspace*{-15pt}
    \caption{\textbf{Dominance Values}: We compare the dominance values obtained using the user study and those predicted by our PDM model for some of the scenarios.}
  \label{fig:dominanceValues}
    \vspace*{-20pt}
    \end{figure}

The values of the coefficients ($\textbf{D}$) indicate how the parameters of the motion model affect the perceived dominance levels of pedestrians. Positive values of coefficients for Preferred~Speed and Radius indicate that dominant pedestrians generally walk faster and have larger interpersonal distances. A large negative coefficient for Planning~Horizon indicates that pedestrians who do not plan far ahead are perceived to be more dominant. A negative coefficient for Maximum~Neighbors indicates that a pedestrian with a lower number of Maximum Neighbors is more likely to be perceived as dominant rather than submissive. These results comply with the findings of psychological research on dominance, which say that dominant agents tend to act aggressively~\cite{mcdermott2017dominance} and expect others to accommodate and acquiesce to their behavior~\cite{cohen1996insult}. A positive value for the coefficient of Neighbor~Distance is not supported by the psychological literature, but the value is negligibly small.

We compare the dominance values obtained from the user ($d_i^{user}$) and the values predicted by the PDM model ($d_i^{PDM}$). We show a sample of $24$ values in Figure~\ref{fig:dominanceValues}. We compute the error between the two values as:
\begin{equation}
e(d_i^{user}, d_i^{PDM}) = |d_i^{user} - d_i^{PDM}|
\end{equation}
We compute the average error for all $48$ data points using leave-one-out cross-validation. We observe an average error of $0.15$ between the predicted and user-obtained dominance values, i.e. our PDM can predict dominance from motion model parameters with \texttildelow 85\% accuracy. 

\section{Socially-Aware Robot Navigation}
Psychological research shows that, in an interaction, humans who exhibit complementing responses (dominance in response to submission and submission in response to dominance) are more comfortable with their interaction partner and like their partner more~\cite{tiedens2003power}. We use this finding to provide socially-aware robot navigation such that the robot exhibits dominance behaviors that are complementary to the pedestrians' dominance levels. We explain the details of this algorithm in this section.

We assume that the environment contains $n$ pedestrians and $m$ robots. We use Reciprocal Velocity Obstacles (RVO)~\cite{van2011reciprocal} for collision-free navigation for each robot. It is expected that the pedestrians move towards their goals while trying to avoid collisions with each other and with robots. Our approach can also be combined with other motion models based on Boids or social forces, instead of RVO. We extract trajectories from real-world video input. We compute the pedestrian motion model (RVO) parameters using Bayesian learning, which also compensates for noise and incomplete trajectories~\cite{bera2017sociosense}.

We compute the dominance $d_{i}$ of each pedestrian ($i=1...n$) with the help of PDM (Equation~\ref{eq:mapping}). For a single pedestrian with dominance $d_1$, the desired dominance value of each robot in the group will be $(1 - d_1)$ because we want the robots to exhibit complementary behavior. For multiple pedestrians each with a different dominance value $d_i$, our desired robot dominance value $d_{des}$ is computed as a solution to the following optimization:
\begin{eqnarray}
    \underset{d}{\text{minimize}} \sum_{i=1}^{n} (d - (1 - d_i))^2
\end{eqnarray}

This equation tries to compute the dominance scalar that minimizes the difference from the ideal complementary dominance values $(1-d_i)$ for each pedestrian. This optimization has the solution:

\begin{eqnarray}
    d_{des} = \frac{\sum_{i=1}^{n} (1-d_i)}{n}
\end{eqnarray}

Based on Equation~\ref{eq:mapping} and Equation~\ref{eq:coeffs}, there can be multiple sets of motion model parameters that have the desired dominance values for the robots. Depending on the mechanical constraints of the robots, we can choose the parameters $\mathbf{P}_{des}$ that minimize the cost of navigation in the environment. Suppose the cost of navigation for a robot as a function of normalized motion model parameters $\mathbf{P}$ is represented as $c_{robot}(\mathbf{P})$. Next, we compute $\mathbf{P}_{des}$ by minimizing:
\begin{eqnarray}
    & \underset{\mathbf{P}}{\text{minimize}} & c_{robot}(\mathbf{P}),  \\
    & \text{subject to } & d_{des} = \mathbf{D}* \mathbf{P},
\end{eqnarray}
where $\mathbf{D}$ is the coefficient vector from Equation~\ref{eq:coeffs}.

\subsection{Performance Evaluation}
To analyze the performance of our navigation algorithm, we extracted the trajectories of pedestrians in different benchmark videos. These crowd videos contain videos with low pedestrian density (less than 1 pedestrian per square meter), medium pedestrian density (1-2 pedestrians per square meter), and high pedestrian density (more than 2 pedestrians per square meter). We obtained the timing results (Table~\ref{tab:dataset}) for motion model parameter computation and complementary dominance computation on a Windows 10 desktop PC with Intel Xeon E5-1620 v3 with 16 GB of memory, which uses four cores.

\begin{table}[ht]
\begin{center}
\scalebox{0.8}{
\begin{tabular}{|l|l|l|l|l|l|l|l|}
\hline
\textbf{Scenario}                  & \ \textbf{Analysed} & \ \textbf{Input} &\ \  \textbf{Avg. time }    &\ \textbf{Avg. time} \\
                                  & \textbf{Pedestrians}      & \textbf{Frames}   & \textbf{Motion Model}         &\textbf{Dominance}\\ \hline\hline
Manko              & 42         & 373      & 0.034         &  93E-06\\ \hline
Marathon            & 18         & 450      & 0.041          & 44E-06 \\ \hline
Explosion           & 19         & 238      & 0.033          & 53E-06 \\ \hline
Street             & 147        & 9014     & 0.022         & 280E-06 \\ \hline
TrainStation   & 200            & 999      & 0.061          & 315E-06 \\ \hline
ATC    Mall        & 50           & 7199     & 0.037         & 106E-06 \\ \hline
IITF-1               & 18         & 450      & 0.041          & 44E-06 \\ \hline
IITF-3        & 19         & 238      & 0.046          & 54E-06 \\ \hline
IITF-5           & 18         & 450      & 0.056          & 44E-06 \\ \hline
NPLC-1        & 19         & 238      & 0.039          & 53E-06 \\ \hline
NPLC-3        & 18         & 450      & 0.031          & 52E-06 \\ \hline
\end{tabular}}
\end{center}
\caption{\textbf{Performance on the Benchmarks:} We present the performance of motion model computation and dominance identification algorithms on different benchmark videos. We highlight the number of video frames of extracted trajectories, the number of pedestrians used for dominance identification, and the computation time (in seconds).}
\vspace*{-15pt}
\label{tab:dataset}
\end{table}

\section{Vehicle-Pedestrian Interaction}
PDM can also be used to predict pedestrian behavior in scenarios involving interactions between autonomous vehicles and pedestrians. In this section, we apply dominance identification to a vehicle-pedestrian interaction scenario. Dominant people often ignore norms and disregard oncoming physical threats to seek more power within social environments~\cite{cohen1996insult}. We consider these findings from Psychology and use pedestrian dominance to facilitate socially-aware interactions between autonomous vehicles and pedestrians in shared spaces.

\begin{figure}
  \centering
  \includegraphics[width =1.0 \linewidth]{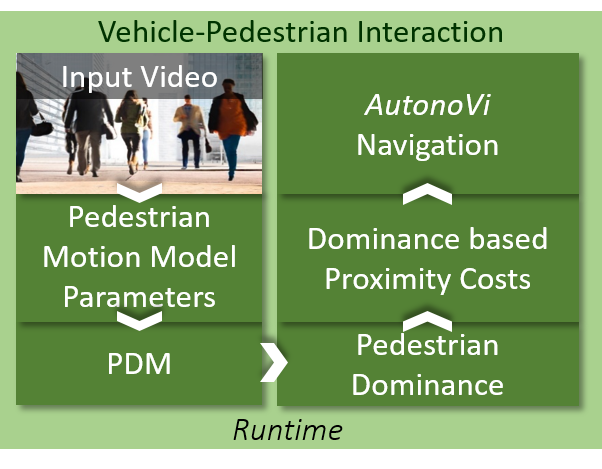}
    \vspace*{-15pt}
    \caption{\textbf{Autonomous Vehicle Navigation:} We apply PDM to vehicle-pedestrian interaction scenarios. From an input video, we extract the motion model parameters of pedestrians using Bayesian learning. We then use PDM to identify the dominance level of each pedestrian. We make use of the AutonoVi algorithm~\cite{best2017autonovi} for navigation. We update the proximity cost computation for AutonoVi using personalized dominance-based proximity costs for each pedestrian. Our algorithm thus facilitates dominance-aware interactions between autonomous vehicles and pedestrians.}
  \label{fig:applicationOverview}
    \vspace*{-10pt}
    \end{figure}

\begin{figure}
  \centering
  \includegraphics[width =1.0 \linewidth]{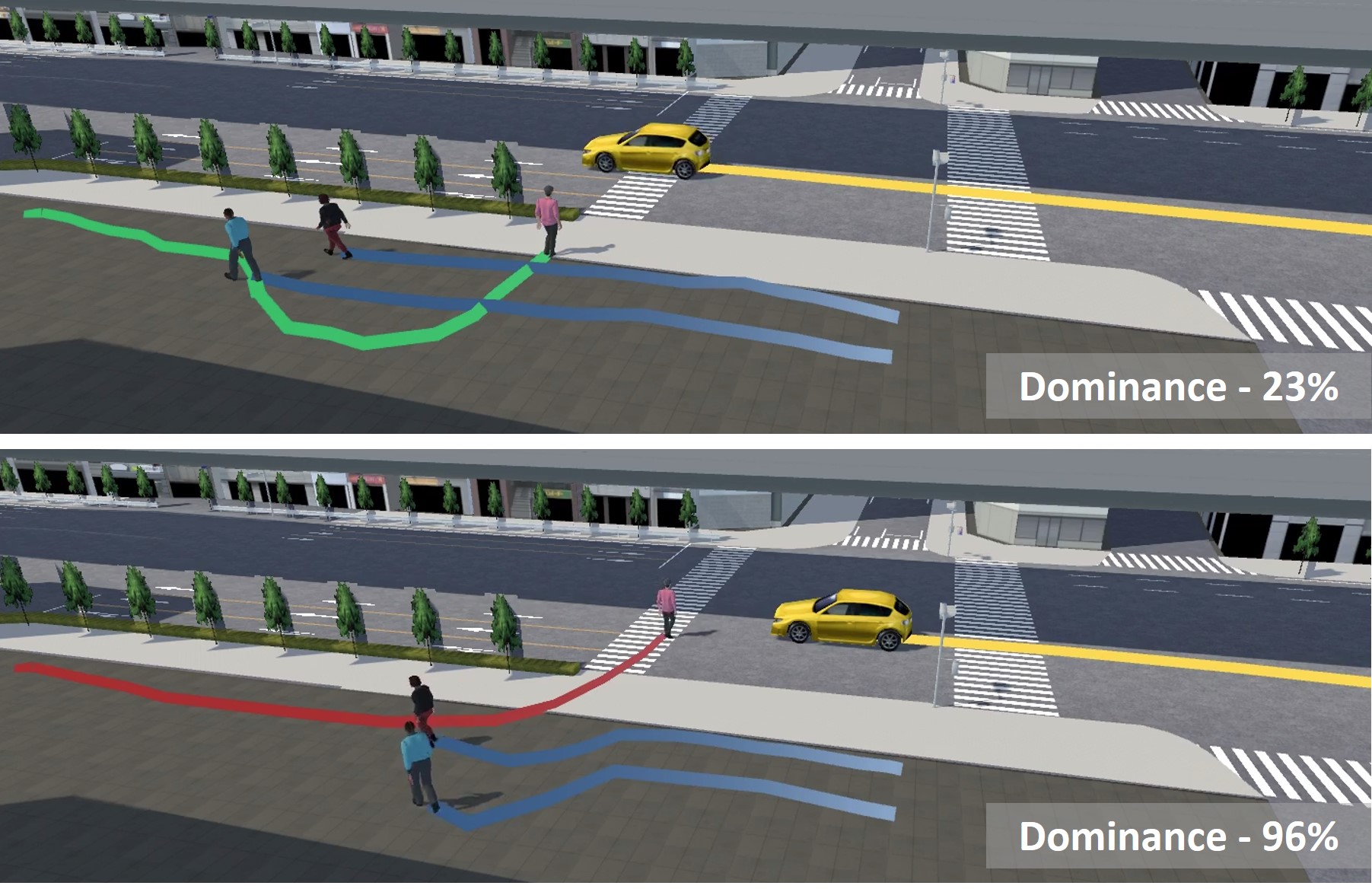}
    \vspace*{-15pt}
    \caption{\textbf{Vehicle-Pedestrian Interaction}: The vehicle makes a navigation decision based on the pedestrian dominance. If PDM identifies the pedestrian (\textbf{\textcolor{green}{green}} trajectory) as submissive (top image), the vehicle continues to navigate on its current path (marked by \textbf{\textcolor{orange}{yellow}}) and predicts that the pedestrian will stop to let it pass. If PDM identifies a pedestrian (\textbf{\textcolor{red}{red}} trajectory) as dominant (bottom image), the vehicle stops and lets the pedestrian pass.}
  \label{fig:application}
    \vspace*{-20pt}
    \end{figure}

Figure~\ref{fig:applicationOverview} provides an overview of our algorithm. Our algorithm uses AutonoVi~\cite{best2017autonovi} for vehicle navigation simulation. AutonoVi is based on minimizing a cost function that includes path, comfort, maneuverability, and proximity costs. When the vehicle is located at position $p_v$, the proximity cost $c_{prox}(p_{i}, p_{v})$ for a pedestrian $i$ with position $p_i$ is computed as follows:
\begin{eqnarray}
    c_{prox}(p_{i}, p_{v}) = C_{ped} e^{-\norm{p_i - p_v}}
\end{eqnarray}
Here, $C_{ped}$ is a constant coefficient for pedestrian neighbors. AutonoVi uses the same value of $C_{ped}$ for all pedestrians and does not consider the varying personalities of pedestrians. We model the pedestrian coefficient for a pedestrian $i$ as a function of their dominance $d_i$:
\begin{eqnarray}
    C_{ped}(i) = C_{ped}e^{-s \cdot d_i}
\end{eqnarray}
where $s \in [0, 1]$ is a safety variable that controls the usage of dominance prediction in vehicle navigation. A value of $s=0$ corresponds to $C_{ped}(i) = C_{ped} \forall i$, i.e. dominance will not be used for vehicle navigation. The value of $s$ can be set according to the vehicle's dynamic variables, e.g., speed, acceleration, etc. 

The proximity cost for the pedestrian $i$ then becomes:
\begin{eqnarray}
    c_{prox}(p_{i}, p_{v}) = C_{ped} e^{-s \cdot d_i-\norm{p_i - p_v}}
\end{eqnarray}
Our navigation algorithm then uses these pedestrian-specific proximity costs in AuonoVi's cost function.

Figure~\ref{fig:application} shows a vehicle-pedestrian interaction scenario where the vehicle makes a navigation decision based on the pedestrian's dominance level. When the vehicle and the pedestrians have to share the same space, the vehicle uses PDM to identify the pedestrian dominance. Since dominant people expect others to move out of their way~\cite{ridgeway1987nonverbal}, in the case of a  dominant pedestrian, the vehicle stops to let the pedestrian pass. If PDM identifies the pedestrian as submissive, the vehicle can continue to navigate on its current path.
\section{Conclusion, Limitations, and Future Work}
We present a novel Pedestrian Dominance Model for identifying the dominance of a pedestrian based on his or her trajectory information. We perform a user study to obtain perceived dominance values for pedestrians in a dataset of simulated videos. Using multiple linear regression, we establish a linear mapping between dominance values and pedestrians' motion model parameters. We present an algorithm for identifying pedestrian dominance in real-time from input videos using Bayesian learning. Our formulation for socially-aware robot navigation is based on prior work in psychology literature, which states that complementarity in dominance increases the rapport and comfort between two interacting partners (humans and/or robots). We compute motion model parameters that generate dominance behaviors for robots that will be complementary to those of pedestrians. We also apply pedestrian dominance computation to the navigation of autonomous vehicles around pedestrians. 

There are some limitations to our approach. Our model currently considers all the pedestrians around the robot while calculating the motion model parameters. In future, we would like to assign more importance to pedestrians in the robot's field of vision and/or immediate vicinity. We consider trajectory information while evaluating the dominance of pedestrians, but humans also exhibit other non-verbal and verbal dominance cues. We would like to consider gait, gestures, gazing, and facial expressions when making judgments about the dominance levels of pedestrians. We would also like the robots to complement these non-verbal indicators of dominance. Our socially-aware robot navigation uses complementary behaviors to increase the rapport and comfort between the robots and pedestrians, but there are situations where the robots must perform different tasks such as crowd control. We would like to investigate these situations and design appropriate robot behaviors for them.

\section*{ACKNOWLEDGMENT}
This research is supported in part by ARO grant W911NF-19-1- 0069, Alibaba Innovative Research (AIR) program and Intel.

\bibliographystyle{IEEEtran}
\bibliography{ref}
\end{document}